\documentclass[runningheads]{llncs}
\usepackage{graphicx}

\usepackage{tikz}
\usepackage{comment}
\usepackage{amsmath,amssymb} 
\usepackage{color}
\usepackage{bm}

\usepackage{algorithm}
\usepackage{algorithmic}
\usepackage{multirow}
\usepackage{mathrsfs}
\usepackage{makecell}
\usepackage{booktabs}
\usepackage{bbm}
\usepackage{graphicx}
\usepackage{subfigure}
\usepackage{enumitem}
\usepackage{wrapfig}
\usepackage{url}
\usepackage{lipsum}
\usepackage[symbol]{footmisc}
\renewcommand{\thefootnote}{\fnsymbol{footnote}} 
\usepackage[colorlinks=true,citecolor=green]{hyperref}

\usepackage[accsupp]{axessibility}  

\newcommand{\etal}{{\em et al\,.}}       
\newcommand{\eg}{{\em e.g.}}           
\newcommand{\ie}{{\em i.e.}}           
\newcommand{\etc}{{\em etc.}}         
\newcommand{\vs}{{\em vs. }}           

\begin{document}
\begin{sloppypar}
\pagestyle{headings}
\mainmatter
\def\ECCVSubNumber{482}  

\title{Generalizable Medical Image Segmentation via Random Amplitude Mixup and Domain-Specific Image Restoration} 

\titlerunning{RAM-DSIR}
%
\author{Ziqi Zhou\inst{1, 2} \and
Lei Qi\inst{3}\textsuperscript{*} \and
Yinghuan Shi\inst{1, 2}\textsuperscript{*}}
\authorrunning{Zhou et al.}
%
\institute{State Key Laboratory for Novel Software Technology, Nanjing University \and
National Institute of Healthcare Data Science, Nanjing University \and
School of Computer Science and Engineering, Southeast University \email{zhouzq@smail.nju.edu.cn, qilei@seu.edu.cn, syh@nju.edu.cn}}
\maketitle
\newcommand\blfootnote[1]{%
\begingroup
\renewcommand\thefootnote{}\footnote{#1}%
\addtocounter{footnote}{-1}%
\endgroup
}
\begin{abstract}
For medical image analysis, segmentation models trained on one or several domains lack generalization ability to unseen domains due to discrepancies between different data acquisition policies. We argue that the degeneration in segmentation performance is mainly attributed to overfitting to source domains and domain shift. To this end, we present a novel generalizable medical image segmentation method. To be specific, we design our approach as a multi-task paradigm by combining the segmentation model with a self-supervision \textit{domain-specific image restoration} (DSIR) module for model regularization. We also design a \textit{random amplitude mixup} (RAM) module, which incorporates low-level frequency information of different domain images to synthesize new images. To guide our model be resistant to domain shift, we introduce a semantic consistency loss. We demonstrate the performance of our method on two public generalizable segmentation benchmarks in medical images, which validates our method could achieve the state-of-the-art performance. \textcolor{red}{\textsuperscript{$\ddagger$}}
\blfootnote{\textsuperscript{*}Corresponding Authors: Yinghuan Shi and Lei Qi.}
\blfootnote{\textsuperscript{$\ddagger$}Code is available at \href{https://github.com/zzzqzhou/RAM-DSIR}{https://github.com/zzzqzhou/RAM-DSIR}.}

\end{abstract}

\section{Introduction}
Recently, deep convolution neural networks (DCNNs) have progressed remarkably in computer vision tasks (\eg, image classification, semantic segmentation, object detection, \etc). Especially in medical image segmentation tasks, deep learning based methods have taken over the dominant position~\cite{ronneberger2015u,milletari2016v}. Usually, DCNNs require large numbers of annotated training images to alleviate the risk of overfitting. However, datasets in medical image segmentation tasks are often relatively small in amount than those in natural image segmentation tasks. Moreover, it is notoriously time-consuming to acquire segmentation annotations of medical images. Accurate annotations also requires specific expertise in radiodiagnosis. Except for the data amounts and annotations problem, basic deep learning methods assume that training data and test data share same distribution information. This assumption requires that training and test data are collected from the same distribution, which is a strong assumption. Due to data distribution shifts, this assumption usually becomes invalid in the real clinical setting. It is known that the quality of medical images varies greatly due to many factors, such as different scanners, imaging protocols, and operators. As a result, the segmentation model directly trained on a set of training images may lack generalization ability on test images drawn from another hospital or medical center, which follows a different distribution.

To fight against distribution shift, tremendous researchers have investigated several practical settings, such as unsupervised domain adaptation (UDA), domain generalization (DG), \etc ~UDA-based segmentation methods have gained much popularity in medical image segmentation~\cite{hoffman2018cycada,chen2019crdoco,zhang2018fully}. To be specific, UDA attempts to learn a segmentation network on single or multiple source domain images including the annotations along with unlabeled target domain images. UDA-based methods intend to narrow the domain gap between source and target domains. However, this prerequisite is sometimes impractical or infeasible in real-world application. Since data privacy protection is rigorous in medical image scenarios, we may sometimes have no chance to access target domain images from some medical centers.

\begin{figure}[t]
    \centering
    \includegraphics[width=0.95\textwidth]{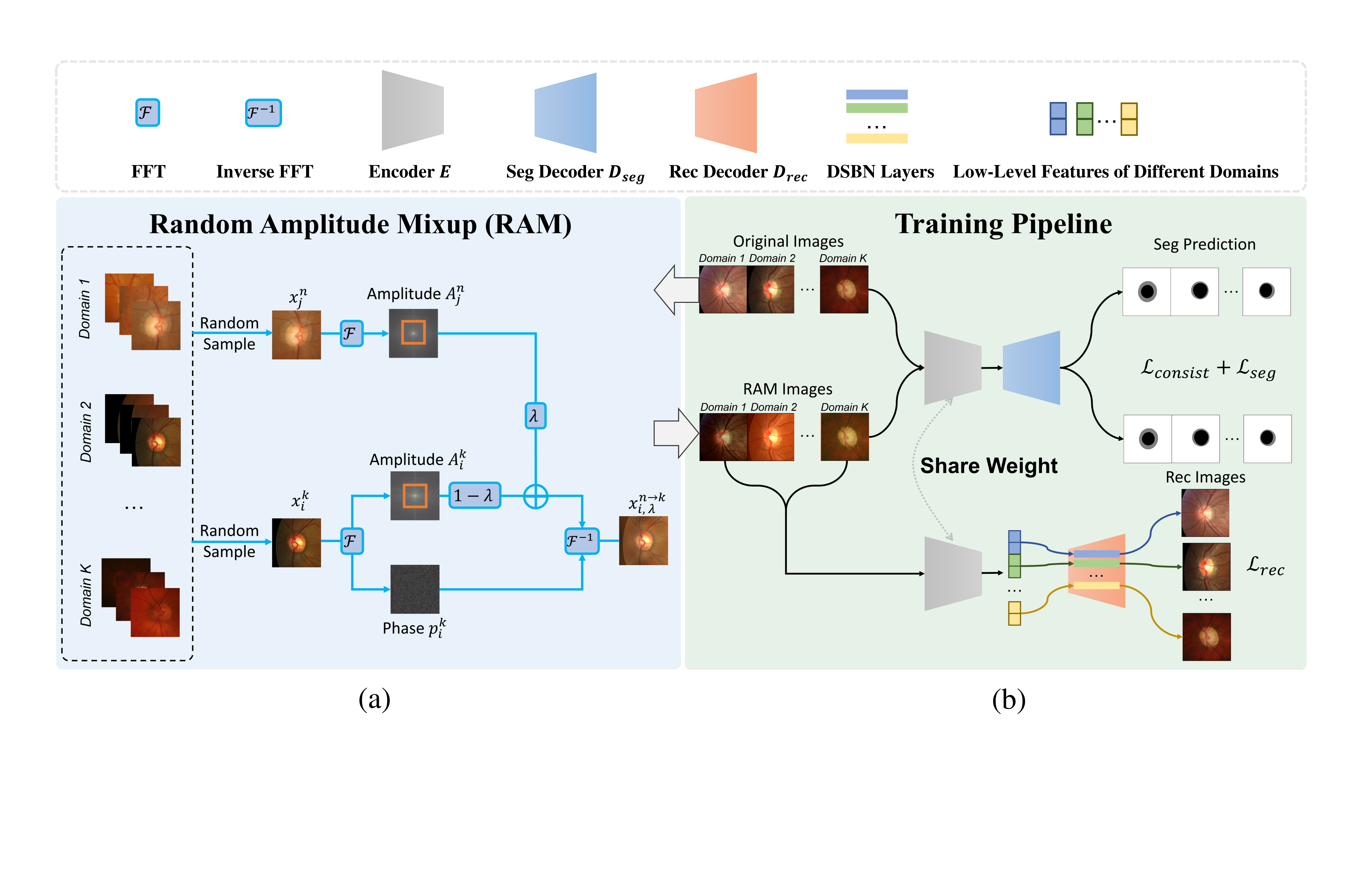}
    \caption{The overall architecture of our method. (a) Random Amplitude Mixup (RAM): We extract the amplitude maps from two random sampled images of different domains and incorporate the amplitude maps of them. Then we can synthesize new images that have different domain styles and preserve original semantic information. (b) The synthesized images from RAM module are utilized to train the segmentation model and DSIR decoder. Basic segmentation loss combined with semantic consistency loss and image recovering loss are employed to train our network.}
    \label{architecture}
\end{figure}

Recently, domain generalization (DG) is proposed to alleviate the application limitation in UDA. DG is a more feasible yet challenging setting requiring only source domains for training. After training on source domain images, we can directly deploy the segmentation model to new unseen target domains. Recently, several literature have developed domain generalization methods to improve model generalization ability with multiple source domains. Among these previous methods, most of them attempt to learn a domain-irrelevant feature representation among multi-source domains for generalization~\cite{dou2019domain,li2018domain,li2018learning,li2019episodic}. Some data augmentation based methods have emerged to tackle the problem of lack of prior information from target domains~\cite{yue2019domain,zhang2020generalizing} by synthesizing newly stylized images to expand diversities of source domain images. Some pioneers have proposed self-supervised tasks (\ie, solving  a Jigsaw Puzzles) to help regularize the model~\cite{carlucci2019domain,wang2020learning}. These 
methods indicate that an auxiliary self-supervised task can better help the model learn domain-invariant knowledge, thus improving model regularization. However, solving a Jigsaw Puzzles may not be a sufficient self-supervision for DG segmentation tasks. To this end, we aim to design a more complex self-supervision to better learn domain-invariant semantic representation for medical image segmentation.

In our work, we present a new framework based on vanilla generalizable medical image segmentation model. To be specific, we first introduce a \textit{random amplitude mixup} (RAM) module by utilizing the Fourier transform to capture frequency space signals from different source domain images and incorporating low-level frequency information of different source domain images to generate new images with different styles. We then use these synthetic images as data augmentation to train the segmentation model and improve robustness. To further regularize our model and combat domain shifts, we employ a semantic consistency training loss to minimize the discrepancy between predictions of real source domain images and synthetic images. To learn more robust feature representation, we introduce a \textit{domain-specific image restoration} (DSIR) decoder to recover low-level features from synthetic images to original source domain images. We demonstrate the effectiveness of our approach on two DG medical image segmentation benchmarks. Our method achieves the state-of-the-art performance compared with competitive methods. We display the overall architecture of our method in Fig.~\ref{architecture}.

\section{Related Work}
\textbf{Unsupervised Domain Adaptation.} Unsupervised Domain Adaptation (UDA) is a particular branch of Domain Adaptation (DA) that leverages labeled data from one or multiple source domains along with unlabeled data from the target domain to learn a classifier for the target domain~\cite{chang2019all,chang2019domain,chen2019crdoco,hoffman2018cycada,liu2020open,vu2019advent,zhang2018fully}. Under such a problem setting, data from the target domain can be utilized to guide the optimization procedure. The general motivation of UDA is to align the source domain and target domain distributions. Some methods adopted a generative model to narrow the pixel-level distribution gap between source and target domains~\cite{chen2019synergistic,chen2019crdoco,hoffman2018cycada,zhang2018fully}. 
Dou \etal~\cite{dou2018ijcai} aligned the feature distribution between source and target domains by adversarial training to keep semantic features consistent in different domains. Differently, some methods attempted to narrow the distribution gap between source and target domains in output space level~\cite{chen2018road,tsai2018learning,tsai2019domain,vu2019advent}. However, due to data privacy protection, some unlabeled target domain data can not be accessed in some cases. The target domain is not available in the training process, making UDA methods impractical in some real-world applications.

\textbf{Domain Generalization.} In contrast to UDA, Domain Generalization (DG) purely trains a model on one or more related source domains and directly generalizes to target domains. A large amount of DG methods have been proposed recently~\cite{chattopadhyay2020learning,du2020learning,gong2019dlow,hoffer2020augment,qiao2020learning,yue2019domain,zakharov2019deceptionnet}. Some methods tried to minimize the domain discrepancy across multiple source domains to learn domain-invariant representations~\cite{gong2021confidence,hsu2017learning,li2018domain}. With the recent advance of the episodic training strategy for domain generalization~\cite{balaji2018metareg,dou2019domain,li2018learning,li2019episodic}, some meta-learning-based methods have been developed to generalize models to unseen domains. Li \etal~\cite{li2019episodic} proposed an episodic training procedure to simulate domain shift at runtime to improve the robustness of the network. Unlike previous meta-learning-based methods, our method is based on vanilla training policy by aggregating different source domain images. We apply a self-supervised image-level-recovering task and semantic consistency training policy to improve the generalization performance on unseen target domains. In medical image segmentation, several prior literature have studied DG segmentation. For instance, Zhang \etal~\cite{zhang2020generalizing} proposed a deep-stacked transformation approach that utilized a stack of image transformations to simulate domain shift in medical imaging. Liu \etal~\cite{liu2020saml} introduced a shape-aware supervision combined with meta-learning to help generalizable prostate image segmentation. Wang \etal~\cite{wang2020dofe} stored domain-specific prior knowledge in a pool as domain attributes for domain aggregation. Liu \etal~\cite{liu2021feddg} proposed a continuous frequency space augmentation with episodic training policy to improve the generalization ability across different domains. Similar to Liu \etal~\cite{liu2021feddg}, we apply frequency space information for image augmentation in our method. However, we utilize augmented images for image segmentation and the auxiliary image-level-recovering task. This will help our model be more robust to domain shifts and alleviate overfitting.

\textbf{Self-supervisied Regularization.} Self-supervised learning have gained much attention in computer vision, natural language processing \etc~\cite{devlin2018bert,he2020momentum,bao2021beit}, recently. It utilizes annotation-free tasks to learn feature representations of data for the downstream tasks. In DG scenario, some methods have also introduced self-supervision tasks to regularize the semantic feature learning~\cite{carlucci2019domain,wang2020learning}. We also develop an image-level-recovering self-supervision task to help regularize the model. Different from~\cite{carlucci2019domain,wang2020learning} solving a Jigsaw Puzzles, our image-level-recovering task is more complicated, which can better regularize the model.

\section{Our Method}

\subsection{Definition and Overview}
We denote a set of $K$ source domains as $D_s = \{(x_i^k, y_i^k)_{i=1}^{N_k}\}_{k=1}^{K}$, where $x_i^k$ is the $i$-th image from $k$-th source domain; $y_i^k$ is the segmentation label of $x_i^k$; $N_k$ is the number of samples in $k$-th source domain. We aim to learn a generalizable medical image segmentation model $F_\theta$ on $D_s$. The model $F_\theta$ is expected to show a satisfactory generalization performance on unseen target domain $D_t = \{x_i\}_{i=1}^{N_t}$, where $x_i$ represent the $i$-th image in target domain, and $N_t$ is the number of image samples in target domain.

Our proposed method contains an encoder-decoder segmentation model with an auxiliary \textit{domain-specific image restoration} (DSIR) decoder. In front of our training pipeline, we introduce a data augmentation and corruption module named as \textit{random amplitude mixup} (RAM). The workflow of our method contains three steps. First, in the RAM module, we apply the Fourier transform on two source domain images that share different domain labels to obtain their frequency space signals; then, we incorporate their low-frequency signals and utilize inverse Fourier transform to generate new images. Secondly, in our DSIR module, the encoder of the segmentation model obtains low-level features of images generated by RAM. A decoder with domain-specific batch normalization is trained to recover original images in a specific source domain from the low-level features. Finally, the encoder-decoder segmentation model is trained by the segmentation loss of source domain images and augmented images; also we adopt a consistency loss between the outputs of source domain images and augmented images to help the segmentation model better resist domain shifts. We discuss all of these components next in detail.

\subsection{Random Amplitude Mixup}
To address the restriction of domain discrepancy between source and target domains, a reasonable idea is to apply data augmentation on source domains to diversify source domain data. In this case, we can regularize the model and alleviate overfitting to source domains. Among plenty of data augmentation methods, Mixup~\cite{zhang2018mixup} has been widely used in image recognition tasks. Image-level-Mixup (IM) incorporates two different images from the training dataset. However, IM will also disturb the semantic information of images, which may negatively influence semantic segmentation tasks. Inspired by prior literature~\cite{yang2020fda,liu2021feddg}, we propose to exploit the inherent information of source domains in the frequency space and incorporate distribution information (\ie, style) in the amplitude spectrum of different images. We name our module as \textit{random amplitude mixup} (RAM).

To be specific, we randomly take a sample image $x_i^k \in \mathbb{R}^{H \times W \times C}$ ($C$ represents the number of image channels; $H$ and $W$ are height and width of the image) from source domain $k$. Then, we perform the Fourier transform~\cite{nussbaumer1981fast} $\mathcal{F}$ to obtain the frequency space signal of image $x_i^k$, which can be written as:
\begin{equation}
    \mathcal{F}(x_i^k)(u,v,c) = \sum\limits_{h=0}^{H-1}\sum\limits_{w=0}^{W-1}x_i^k(h,w,c)e^{-j2\pi(\frac{h}{H} u + \frac{w}{W}v)}, j ^ 2 = -1.
\end{equation}
After the Fourier transform, we can decompose the frequency signal $\mathcal{F}(x_i^k)$ into an amplitude spectrum $\mathcal{A}_i^k \in \mathbb{R}^{H \times W \times C}$ and a phase image $\mathcal{P}_i^k \in \mathbb{R}^{H \times W \times C}$, where the amplitude spectrum contains low-level statistics (\eg, style) while the phase image includes high-level (\eg, object) semantics of the original image. We incorporate the amplitude spectrum of different images from multiple source domains. To this end, we randomly select another sample image $x_j^n (n \neq k)$ from source domain $n$ and perform the Fourier transform on it as well. So that, we obtain another amplitude $\mathcal{A}_j^n$ of image $x_j^n$. To incorporate the low-frequency component within amplitude $\mathcal{A}_i^k$ and $\mathcal{A}_j^n$, we introduce a binary mask $\mathcal{M}$ which can control the scale of low-frequency component in amplitude spectrum to be incorporated. After that, we incorporate the amplitude information of image $x_i^k$ and image $x_j^n$ by:
\begin{equation}
    \mathcal{A}_{i, \lambda}^{n \rightarrow k} = \mathcal{A}_{i}^{k} * (1 - \mathcal{M}) + ((1 - \lambda)\mathcal{A}_{i}^{k} + \lambda\mathcal{A}_{j}^{n}) * \mathcal{M},
\end{equation}
where $\mathcal{A}_{i, \lambda}^{n \rightarrow k}$ is the newly interpolated amplitude spectrum; $\lambda$ is a parameter that used to adjust the ratio between $\mathcal{A}_{i}^{k}$ and $\mathcal{A}_{j}^{n}$. Finally, we can transform the merged amplitude $\mathcal{A}_{i, \lambda}^{n \rightarrow k}$ into a newly stylized image through inverse Fourier transform $\mathcal{F}^{-1}$ as follows:
\begin{equation}
    x_{i, \lambda}^{n \rightarrow k} = \mathcal{F}^{-1} (\mathcal{A}_{i, \lambda}^{n \rightarrow k}, \mathcal{P}_i^k),
\end{equation}
where the generated image $x_{i, \lambda}^{n \rightarrow k}$ contains the semantic information of $x_i^k$ and its low-level information (\eg, style) is a mixture of low-level information of $x_i^k$ and $x_j^n$. In our implementation, we follow~\cite{liu2021feddg} to dynamically sample $\lambda$ from $[0.0, 1.0]$ to generate images. Fig.~\ref{architecture} (a) illustrates the overall architecture of RAM.

\begin{wrapfigure}{r}{0.5\textwidth}
    \begin{center}
    \includegraphics[width=0.5\textwidth]{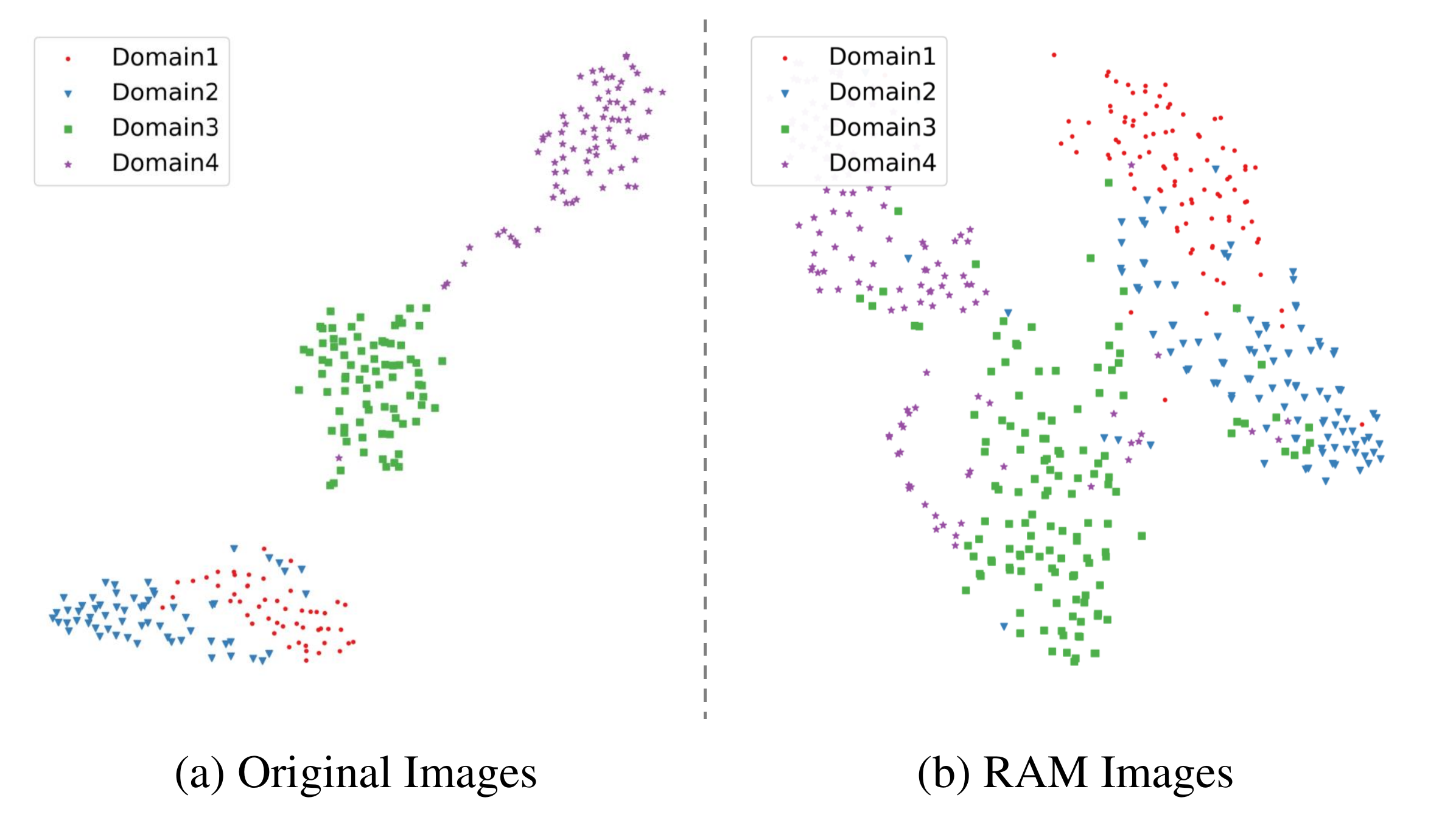}
    \end{center}
    \caption{t-SNE visualization of features of original images and RAM augmented images from \textbf{Fundus} dataset. We use different colors and markers to denote different domains.}
    \label{tsne}
\end{wrapfigure}

To further indicates that RAM can increase the diversity of source domain and narrow the domain discrepancy. We show the t-SNE~\cite{van2008visualizing} visualization of image features in \textbf{Fundus} dataset in Fig.~\ref{tsne}. Fig.~\ref{tsne} (a) shows the original distribution information of different domains in the \textbf{Fundus} dataset. From the visualization, we can observe that the image features from different domains are clearly separated. This leads to the problem that training the model on original source domains make the model easily overfit to specific source domains, which might degrade generalization performance on target domains. However, in Fig.~\ref{tsne} (b), we discover that, by applying RAM on original source domains, we can narrow domain gaps significantly, showing domain invariant representation. The distribution of different domains is more compacted and diversified.

\subsection{Semantic Consistency Training}

To segment images from the target domain, one straightforward method is to train a vanilla segmentation model in a unified fashion by directly feeding multi-source domain images into the model. We name training such a vanilla segmentation model as ``DeepAll''. Although the ``DeepAll'' method might have good generalization performance on multi-source domains, it may not preserve satisfactory segmentation performance on target domain images. Training a vanilla segmentation model on multi-source domains does not introduce supervision to combat domain shifts. Also, we have mentioned that original multi-source domain images lack sufficient diversity in feature distribution, which may lead to overfitting to a specific source domain.

We design a semantic consistency training strategy to tackle problems of the ``DeepAll'' method. To be specific, we introduce an encoder-decoder structure~\cite{ronneberger2015u} as our segmentation model. The encoder $E$ will extract low-level semantic features from images while the segmentation decoder $D_{seg}$ is used to predict segmentation masks. We formulate the forward propagation of the segmentation model on source domain image $x_i^k$ as:
\begin{equation}
    \hat{y}_i^k = D_{seg}(E(x_i^k)),
    \label{forward_1}
\end{equation}
where $\hat{y}_i^k$ is the predicting segmentation mask. Since we utilize RAM to generate newly stylized images from original source domains, we can use these augmented images to help train the segmentation model. This can also regularize the segmentation model and improve its generalization performance on target domains. Similar to Eq.~(\ref{forward_1}), the forward propagation on $x_{i, \lambda}^{n \rightarrow k}$ can be written as:
\begin{equation}
    \hat{y}_{i, \lambda}^{n \rightarrow k} = D_{seg}(E(x_{i, \lambda}^{n \rightarrow k})),
\end{equation}
where $\hat{y}_{i, \lambda}^{n \rightarrow k}$ represents the prediction. Then, we utilize the unified cross-entropy (CE) loss~\cite{murphy2012machine} and Dice loss~\cite{milletari2016v} as our segmentation loss to optimize the model. The CE and dice loss on original source domain $k$ are formulated as:
\begin{equation}
    \mathcal{L}_{ce}^{k} = -\frac{1}{N}\sum_{i=0}^{N-1}\Big(y_i^k\log\hat{y}_i^k + (1 - y_i^k)\log(1 - \hat{y}_i^k)\Big),
\end{equation}
\begin{equation}
    \mathcal{L}_{dice}^{k} = 1 - \frac{2\sum_{i=0}^{N-1} \hat{y}_i^k y_i^k}{\sum_{i=0}^{N-1}(\hat{y}_i^k + y_i^k + \epsilon)},
\end{equation}
where $y_i^k$ is the shared ground truth of $x_i^k$ and $x_{i, \lambda}^{n \rightarrow k}$; $N$ represents the number of samples from domain $k$; $\epsilon$ is a smooth factor to avoid dividing by 0. The CE loss $\mathcal{L}_{ce}^{n \rightarrow k}$ and dice loss $\mathcal{L}_{dice}^{n \rightarrow k}$ on the generated images are similar as above. So, segmentation losses on $x_i^k$ and $x_{i, \lambda}^{n \rightarrow k}$ can be written as:
\begin{equation}
    \mathcal{L}_{seg}^{k} = \mathcal{L}_{dice}^{k} + \mathcal{L}_{ce}^{k},~~~ \mathcal{L}_{seg}^{n \rightarrow k} = \mathcal{L}_{dice}^{n \rightarrow k} + \mathcal{L}_{ce}^{n \rightarrow k}.
\end{equation}

To combat domain shifts, we propose a novel semantic consistency loss in our method. Specifically, we regard the generated image $x_{i, \lambda}^{n \rightarrow k}$ as a style augmentation of $x_i^k$. We intend to force the segmentation model to predict consistent segmentation results from $x_i^k$ and $x_{i, \lambda}^{n \rightarrow k}$. So that the segmentation model can be less sensitive to domain shift. We design a loss term to minimize the Kullback-Leibler (KL) divergence~\cite{kullback1951information} between soft predictions $\hat{y}_i^k$ and $\hat{y}_{i, \lambda}^{n \rightarrow k}$. Our semantic consistency loss is as follows:
\begin{equation}
    \mathcal{L}_{consist}^k = \frac{1}{N}\sum_{i=0}^{N-1} \Big(\text{KL}(\hat{y}_i^k \| \hat{y}_{i, \lambda}^{n \rightarrow k}) + \text{KL}(\hat{y}_{i, \lambda}^{n \rightarrow k} \| \hat{y}_i^k)\Big),
\end{equation}
where KL represents the KL-divergence~\cite{kullback1951information}. We compute a symmetric version of KL-divergence between $\hat{y}_i^k$ and $\hat{y}_{i, \lambda}^{n \rightarrow k}$. By explicitly enhancing the consistency of results, the segmentation model can extract semantic features more robust to domain shift, thus improving performance on unseen target domains.

\subsection{Domain-Specific Image Restoration}
To further regularize the segmentation model and reduce overfitting on source domains, we propose a self-supervised auxiliary task to help train a more robust segmentation model. To be specific, we introduce an image restoration decoder with domain-specific batch normalization (DSBN) layers~\cite{chang2019domain}. The image restoration decoder is utilized to recover image from the low-level features extracted by the segmentation encoder $E$ from the RAM image $x_{i, \lambda}^{n \rightarrow k}$. 

To better recover images of different source domains, we add DSBN in our image restoration decoder. Let our image restoration decoder denote as $D_{rec} = \{D_{rec}^1, D_{rec}^2, \cdots, D_{rec}^K\}$, where $K$ represents the number of source domain, $D_{rec}^k$ is used to recover images from low-level features of RAM images generated by $k$-th source domain images. All of the decoders in $D_{rec}$ share the same model parameters but have different batch normalization layers~\cite{ioffe2015batch}. Since distribution information of multi-source domains is quite different, using different batch normalization layers in different domains can better preserve domain intrinsic features for image restoration. The forward propagation of the image restoration module on source domain $k$ are as follows:
\begin{equation}
    \hat{x}_i^k = D_{rec}^{k}(E(x_{i, \lambda}^{n \rightarrow k})),
\end{equation}
where $E$ is the encoder in our segmentation model; $\hat{x}_i^k$ is the recovering image from $x_{i, \lambda}^{n \rightarrow k}$. We utilize this image restoration decoder as a regularization of the segmentation encoder $E$. We show detailed information of our image restoration module in Fig.~\ref{architecture} (b).

To train the image restoration module, we employ L2 distance as recovering loss to optimize $D_{rec}$ and $E$. The recovering loss on $k$-th source domain are:
\begin{equation}
    \mathcal{L}_{rec}^{k} = \frac{1}{NHWC} \sum\limits_{i=0}^{N-1}\sum\limits_{h=0}^{H-1}\sum\limits_{w=0}^{W-1}\sum\limits_{c=0}^{C-1} \Big(x_i^k(h,w,c) - \hat{x}_i^k(h,w,c)\Big)^2,
\end{equation}
where $N$ represents the number of samples from domain $K$; $H, W, C$ are width, height and channel of the image. 

Overall, we can formulate our whole framework as a multi-task learning paradigm. The total training loss are as follows:
\begin{equation}
    \mathcal{L}_{total} = \frac{1}{K}\sum\limits_{k=1}^{K}\Big(\lambda_1 \mathcal{L}_{seg}^{k} +\lambda_2 \mathcal{L}_{seg}^{n \rightarrow k} + \lambda_3 \mathcal{L}_{rec}^k + \lambda_4 \mathcal{L}_{consist}^k\Big),
    \label{obj_fun}
\end{equation}
where $K$ represents the number of source domains; $\lambda_1$, $\lambda_2$, $\lambda_3$, and $\lambda_4$ are hyper-parameters to balance the weights of basic segmentation loss, consistency loss, and image restoration loss respectively.

\section{Experiments}

\subsection{Datasets}
We evaluate our method on two public DG medical image segmentation datasets as popular used in~\cite{liu2020saml,wang2020dofe,liu2021feddg}: \textbf{Fundus}~\cite{wang2020dofe} and \textbf{Prostate}~\cite{liu2020saml}. The \textbf{Fundus} dataset contains retinal fundus images from 4 different medical centers for optic cup and disc segmentation. Each domain has been split into training and testing sets. For pre-processing, we follow the prior literature~\cite{wang2020dofe} and center-crop disc regions with a $800\times 800$ bounding-box for all of images in \textbf{Fundus} dataset. After that, we randomly resize and crop a $256 \times 256$ region on each cropped images as network input. The \textbf{Prostate} dataset collected T2-weighted MRI prostate images from 6 different data sources for prostate segmentation. All of the images have been cropped to 3D prostate region and 2D slices in axial plane have been resized to $384 \times 384$. For model training, we feed 2D slices of prostate images into our model. We normalize the data individually to [-1, 1] in intensity values on both datasets.

\subsection{Implementation Details}
We employ a UNet-based~\cite{ronneberger2015u} encoder-decoder structure as our segmentation model. The DISR decoder is similar to our segmentation decoder by replacing batch normalization layers with DSBN layers. We implement our experiment with the PyTorch framework on 1 Nvidia RTX 2080Ti GPU with 11 GB
memory. We train our model for 400 epochs on \textbf{Fundus} dataset and 200 epochs on \textbf{Prostate} dataset. For each dataset, we set 8 as training batch size. We also employ the Adam optimizer with
an initial learning rate of 0.001 to optimize our model. To stabilize the training process, the learning rate is decayed by the polynomial rule. Last but not least, we set $\lambda_1$, $\lambda_2$, $\lambda_3$, and $\lambda_4$ as 1, 1, 0.1 and 0.5 empirically in Eq.~(\ref{obj_fun}).

Since the \textbf{Fundus} dataset has already split each domain into training and testing sets, we train our model on training sets of source domains and evaluate on testing sets of target domains. During testing, we first resize $800 \times 800$ test images to size of $256 \times 256$ and get $256 \times 256$ segmentation masks. We then resize segmentation masks to $800 \times 800$ and compute evaluation metrics on them. For \textbf{Prostate} dataset, we directly train segmentation model on source domains and test on target domains. Since original images of \textbf{Prostate} dataset are all 3D volumes, we first get 2D predictions and concatenate all 2D predictions of each 3D sample, then compute evaluation metrics on 3D predictions. When testing on \textbf{Prostate} dataset, we also skip those 2D slices that not contain any prostate region. All of implementations on datasets follow previous methods~\cite{liu2020saml,wang2020dofe}. For evaluation, we adopt commonly-used metric of Dice coefficient (Dice) and Average Surface Distance (ASD) to quantitatively evaluate the segmentation results of whole region and the surface shape respectively. Higher Dice coefficient represents better performance and ASD is the opposite. To avoid randomness, we repeat our experiments for 3 times and report the average performance.

\subsection{Comparison with Other DG methods}

\textbf{Experiment setting.} In our experiments, we follow the practice in prior literature of domain generalization and employ the leave-one-domain-out strategy, \ie, training on $K$ source domains and test on the left one target domain (total $K + 1$ domains). So that, for \textbf{Fundus} and \textbf{Prostate} datasets, we have four and six distinguished tasks, respectively. 

We choose five recent state-of-the-art domain generalization methods to compare with ours and reproduce their results. First of all, the JiGen~\cite{carlucci2019domain} is an effective self-supervised based DG methods for model regularization by solving jigsaw puzzles. The BigAug~\cite{zhang2020generalizing} is an augmentateion based DG method. SAML~\cite{liu2020saml} and FedDG~\cite{liu2021feddg} are two meta-learning based generalizable medical image segmentation methods. Finally, the DoFE~\cite{wang2020dofe} is a domain-invariant feature representation learning approach. We further train a vanilla segmentation model by simply aggregating all source domain images as our baseline model. 

In Tables~\ref{fundus_dice} and \ref{fundus_asd}, we show Dice coefficient and ASD results of different domains in \textbf{Fundus} dataset. All of the methods successfully outperform our baseline method except BigAug~\cite{zhang2020generalizing} (Dice coefficient 85.49\% \vs 85.63\%; ASD 14.18 voxel \vs 13.98 voxel). We assume that this is because BigAug~\cite{zhang2020generalizing} was first designed to augment grey-scale medical images (\eg, CT, MRI, \etc) for domain generalization segmentation tasks. Images in \textbf{Fundus} dataset are all RGB images which have quite different image properties compared with other medical images. So that the generalization performance of BigAug~\cite{zhang2020generalizing} could be degraded. Other methods gain improvements above baseline more or less and prove that different regularization and generalization strategies can help the model to learn more robust feature representation. Compared with these methods, we achieve higher average Dice coefficient and better average ASD on \textbf{Fundus} dataset. This thanks to our RAM and DSIR module. The RAM helps to diversify our source domain images to alleviate overfitting. Also, the image restoration tasks can regularize our model to learn more robust feature representation. Last but not least, we adopt a semantic consistency training policy to resist to domain shift. All of these key components contribute to success of our method on \textbf{Fundus} dataset. Compared with baseline, our method achieves consistent improvements over baseline across all unseen domain settings, with the average performance increase of 3.31\% in Dice coefficient and 3.66 voxel average improvement in ASD.

\begin{table}[t]
    \centering
    \caption{Dice coefficient of different methods on \textbf{Fundus} segmentation task (\%). We mark the top results in \textbf{bold}.}
    \scriptsize
    \setlength{\tabcolsep}{2mm}{
        \begin{tabular}{c|cccc|c}
            \toprule[0.8pt]
            Task & \multicolumn{4}{c|}{Optic Cup/Disc Segmentation} & \multirow{2}{*}{Avg.} \\
            \cmidrule[0.3pt]{1-5}
            Unseen Site & Domain 1 & Domain 2 & Domain 3 & Domain 4 & \\
            \midrule[0.3pt]
            JiGen~\cite{carlucci2019domain} & 82.45/95.03 & 77.05/87.25 & 87.01/\textbf{94.94} & 80.88/91.34 & 86.99 \\
            BigAug~\cite{zhang2020generalizing} & 77.68/93.32 & 75.56/87.54 & 83.33/92.68 & 81.63/92.20 & 85.49 \\
            SAML~\cite{liu2020saml} & 83.72/95.03 & 77.68/87.57 & 84.20/94.49 & 82.08/92.78 & 87.19 \\
            FedDG~\cite{liu2021feddg} & 81.72/95.62 & 77.87/88.71 & 83.96/94.83 & 81.90/93.37 & 87.25 \\
            DoFE~\cite{wang2020dofe} & 84.17/94.96 & \textbf{81.03}/89.29 & 86.54/91.67 & \textbf{87.28}/93.04 & 88.50 \\
            \midrule[0.3pt]
            Baseline & 81.44/95.52 & 77.20/87.96 & 85.11/94.56 & 72.30/90.97 & 85.63 \\
            Ours & \textbf{85.48}/\textbf{95.75} & 78.82/\textbf{89.43} & \textbf{87.44}/94.67 & 85.84/\textbf{94.10} & \textbf{88.94} \\
            \bottomrule[0.8pt]
        \end{tabular}
    }
    \label{fundus_dice}
\end{table}

\begin{figure}[t]
    \centering
    \includegraphics[width=0.90\textwidth]{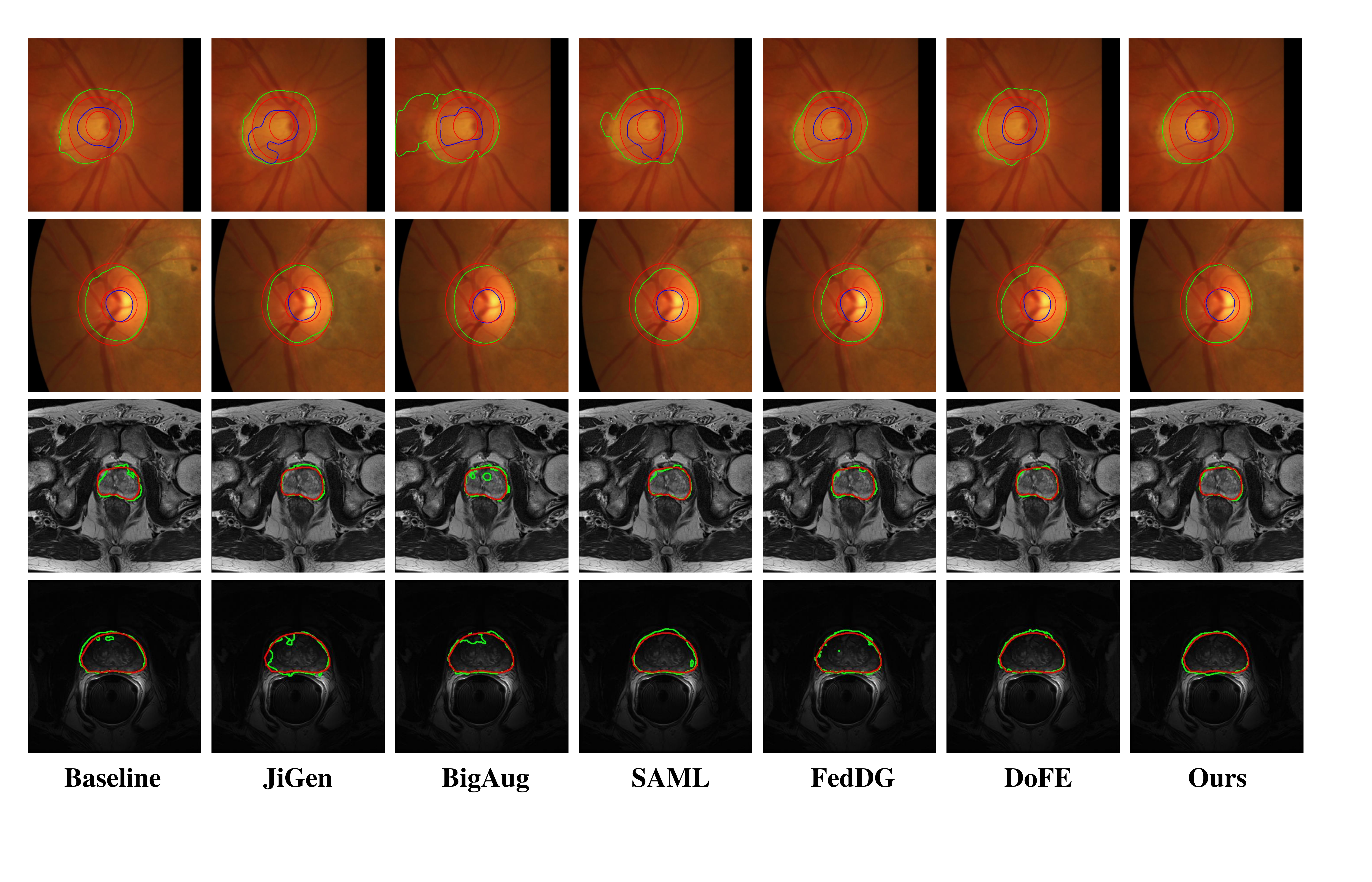}
    \caption{Visualization on segmentation results of different methods on \textbf{Fundus} (top two rows) and \textbf{Prostate} datasets (bottom two rows). The red contours indicate the boundaries of ground truths while the green and blue contours are predictions.}
    \label{vis}
\end{figure}

\begin{table}[t]
    \centering
    \caption{Average Surface Distance (ASD) of different methods on \textbf{Fundus} segmentation task (voxel). We mark the top results in \textbf{bold}.}
    \scriptsize
    \setlength{\tabcolsep}{2mm}{
        \begin{tabular}{c|cccc|c}
            \toprule[0.8pt]
            Task & \multicolumn{4}{c|}{Optic Cup/Disc Segmentation} & \multirow{2}{*}{Avg.} \\
            \cmidrule[0.3pt]{1-5}
            Unseen Site & Domain 1 & Domain 2 & Domain 3 & Domain 4 & \\
            \midrule[0.3pt]
            JiGen~\cite{carlucci2019domain} & 18.57/9.43 & 17.29/19.53 & 9.15/\textbf{6.99} & 15.84/12.14 & 13.62 \\
            BigAug~\cite{zhang2020generalizing} & 22.61/12.53 & 17.95/17.64 & 11.48/10.33 & 11.57/9.36 & 14.18 \\
            SAML~\cite{liu2020saml} & 17.08/9.01 & 16.72/18.63 & 10.87/7.87 & 16.28/8.64 & 13.14 \\
            FedDG~\cite{liu2021feddg} & 18.57/7.69 & 15.87/16.93 & 11.09/7.28 & 10.23/7.51 & 11.90 \\
            DoFE~\cite{wang2020dofe} & 16.07/7.18 & \textbf{13.44}/17.06 & 10.12/10.75 & \textbf{8.14}/7.29 & 11.26 \\
            \midrule[0.3pt]
            Baseline & 18.16/8.99 & 15.67/17.95 & 11.96/9.42 & 20.03/9.64 & 13.98 \\
            Ours & \textbf{16.05}/\textbf{7.12} & 14.01/\textbf{13.86} & \textbf{9.02}/7.11 & 8.29/\textbf{7.06} & \textbf{10.32} \\
            \bottomrule[0.8pt]
        \end{tabular}
    }
    \label{fundus_asd}
\end{table}

\begin{table}[t]
    \centering
    \caption{Dice coefficient of different methods on \textbf{Prostate} segmentation task (\%). We mark the top results in \textbf{bold}.}
    \scriptsize
    \setlength{\tabcolsep}{1mm}{
        \begin{tabular}{c|cccccc|c}
            \toprule[0.8pt]
            Task & \multicolumn{6}{c|}{Prostate Segmentation} & \multirow{2}{*}{Avg.}  \\
            \cmidrule[0.3pt]{1-7}
            Unseen Site & Domain 1 & Domain 2 & Domain 3 & Domain 4 & Domain 5 & Domain 6 & \\ 
            \midrule[0.3pt]
            JiGen~\cite{carlucci2019domain} & 85.45 & 89.26 & 85.92 & 87.45 & 86.18 & 83.08 & 86.22 \\
            BigAug~\cite{zhang2020generalizing} & 85.73 & 89.34 & 84.49 & 88.02 & 81.95 & 87.63 & 86.19 \\
            SAML~\cite{liu2020saml} & 86.35 & 90.18 & 85.03 & 88.20 & 86.97 & 87.69 & 87.40 \\
            FedDG~\cite{liu2021feddg} & 86.43 & 89.59 & 85.30 & 88.95 & 85.93 & 87.39 & 87.27  \\
            DoFE~\cite{wang2020dofe} & \textbf{89.64} & 87.56 & 85.08 & \textbf{89.06} & 86.15 & 87.03 & 87.42 \\
            \midrule[0.3pt]
            Baseline & 85.30 & 87.56 & 82.33 & 87.37 & 80.49 & 81.40 & 84.04 \\
            Ours & 87.56 & \textbf{90.20} & \textbf{86.92} & 88.72 & \textbf{87.17} & \textbf{87.93} & \textbf{88.08} \\
            \bottomrule[0.8pt]
        \end{tabular}
    }
    \label{prostate_dice}
\end{table}

\begin{table}[t]
    \centering
    \caption{Average Surface Distance (ASD) of different methods on \textbf{Prostate} segmentation task (voxel). We mark the top results in \textbf{bold}.}
    \scriptsize
    \setlength{\tabcolsep}{1mm}{
        \begin{tabular}{c|cccccc|c}
            \toprule[0.8pt]
            Task & \multicolumn{6}{c|}{Prostate Segmentation} & \multirow{2}{*}{Avg.}  \\
            \cmidrule[0.3pt]{1-7}
            Unseen Site & Domain 1 & Domain 2 & Domain 3 & Domain 4 & Domain 5 & Domain 6 & \\ 
            \midrule[0.3pt]
            JiGen~\cite{carlucci2019domain} & 1.11 & 1.81 & 2.61 & 1.66 & \textbf{1.71} & 2.43 & 1.89 \\
            BigAug~\cite{zhang2020generalizing} & 1.13 & 1.78 & 4.01 & 1.25 & 1.92 & 1.89 & 2.00 \\
            SAML~\cite{liu2020saml} & 1.09 & 1.54 & 2.52 & 1.41 & 2.01 & 1.77 & 1.72 \\
            FedDG~\cite{liu2021feddg} & 1.30 & 1.67 & 2.36 & 1.37 & 2.19 & 1.94 & 1.81 \\
            DoFE~\cite{wang2020dofe} & \textbf{0.92} & 1.49 & 2.74 & 1.46 & 1.89 & 1.53 & 1.68 \\
            \midrule[0.3pt]
            Baseline & 1.22 & 1.95 & 4.68 & 1.51 & 3.95 & 4.23 & 2.92 \\
            Ours  & 1.04 & \textbf{0.81} & \textbf{2.23} & \textbf{1.16} & 1.81 & \textbf{1.15} & \textbf{1.37} \\
            \bottomrule[0.8pt]
        \end{tabular}
    }
    \label{prostate_asd}
\end{table}

To further indicate effectiveness of our method, we provide experiment results on \textbf{Prostate} dataset in Tables~\ref{prostate_dice} and \ref{prostate_asd}. For prostate segmentation task, all of the comparison DG method outperform baseline. Our method also obtains the highest Dice coefficient and ASD across most unseen domains. The average Dice coefficient 88.08\% and ASD 1.37 voxel are the best compared with other DG methods. Specially, compared with baseline, the increase in overall Dice coefficient of our method is 4.04\% and ASD decreases 1.55 voxel. In Fig.~\ref{vis}, we show the visualization results of two sample images from target domains of \textbf{Fundus} and \textbf{Prostate} datasets. It is explicit that our method can accurately segment the objective structure of unseen domain images and the boundary of the structure is smoother while other methods may fail to do so.

\subsection{Analysis of Our Method}

\begin{figure}[t]
    \centering
    \includegraphics[width=0.90\textwidth]{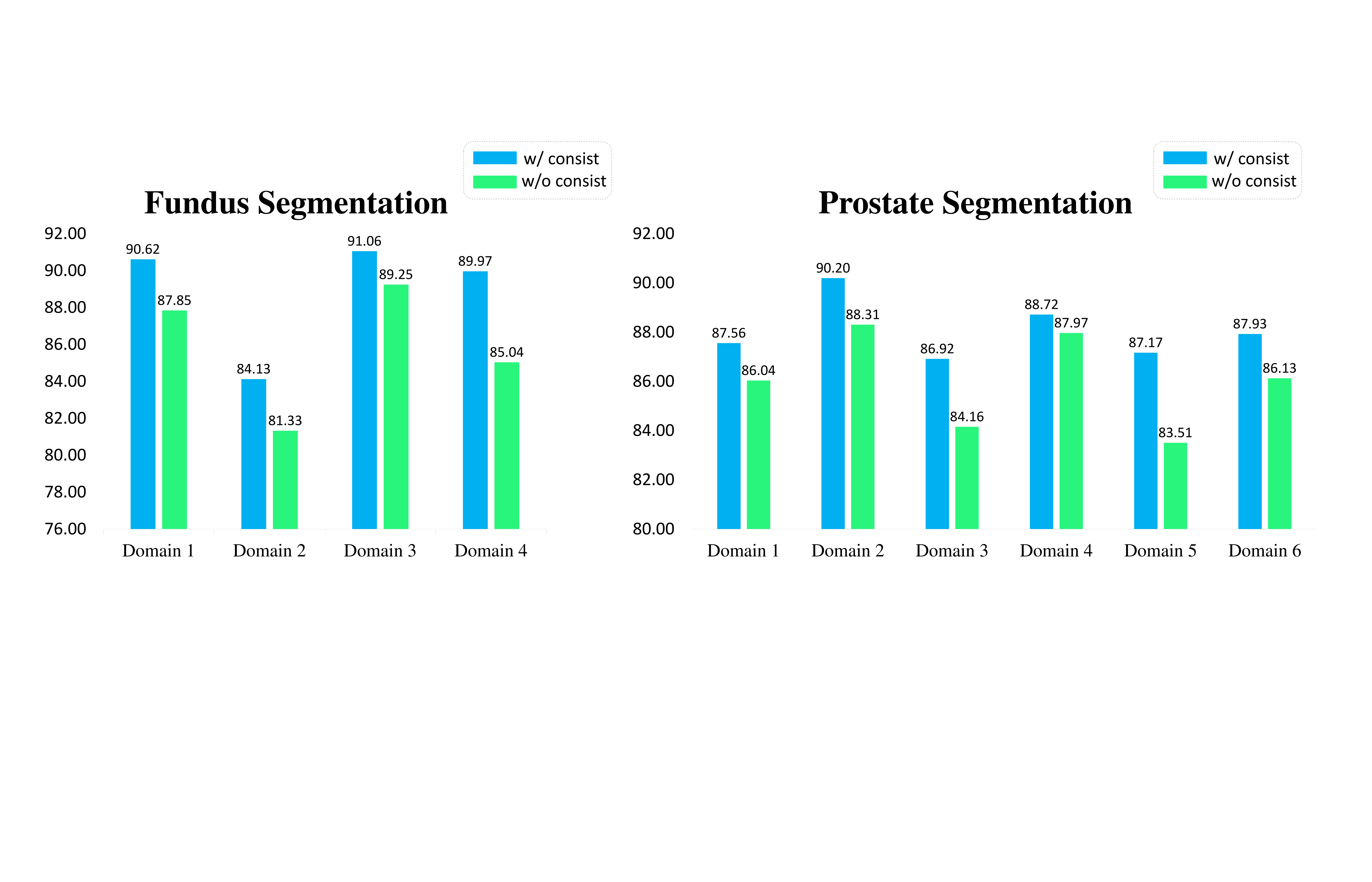}
    \caption{Ablation study of our semantic consistency training policy. Green and blue bars represent average Dice coefficient of our complete method and the method without consistency loss respectively. We show results on different domains from \textbf{Fundus} and \textbf{Prostate} datasets.}
    \label{ablation_consist}
\end{figure}

We conduct extensive ablation studies on our method. Firstly,s we investigate the effectiveness of our random amplitude mixup for data augmentation and DSIR module on \textbf{Fundus} and \textbf{Prostate} dataset. We need to note that, without RAM, our DISR module cannot be implemented. Since our RAM module is utilized to conduct style augmentation and image corruption at the same time, here we discuss the style augmentation and image corruption separately. The experimental results are illustrated in Tables~\ref{fundus_ablation} and \ref{prostate_ablation}. The $\text{RAM}_{\text{Aug}}$ indicates that the RAM style augmentation is employed in our method and DSIR represents the domain-specific image restoration module with image corruption. The method without these two components (\ie, the first row in Tables~\ref{fundus_ablation} and \ref{prostate_ablation}) is the baseline method, which is the same with the baseline results in Tables~\ref{fundus_dice} and \ref{prostate_dice}. From Tables~\ref{fundus_ablation} and \ref{prostate_ablation}, we observe that each component plays a significant role in our method. By adding RAM style augmentation in our method, the overall segmentation performance on fundus segmentation task can increase 2.12\% in Dice coefficient and on prostate segmentation tasks the improvements of Dice coefficient is 3.23\%. Besides, when equipping with domain-specific image restoration module, our model can gain 1.58\% and 1.03\% overall improvements in Dice coefficient on \textbf{Fundus} and \textbf{Prostate} datasets respectively. Based on these results, we justify that, our RAM and DSIR module can help regularize our segmentation model and improve the generalization ability. Last rows in Tables~\ref{fundus_ablation} and \ref{prostate_ablation} display the results by adding all of the components in our method, which are the same as results of our method in Tables~\ref{fundus_dice} and \ref{prostate_dice}. 

\begin{table}[t]
    \centering
    \caption{Ablation Study of key components in our method on \textbf{Fundus} Segmentation Task (\%). We mark the top results in \textbf{bold}.}
    \scriptsize
    \setlength{\tabcolsep}{1mm}{
        \begin{tabular}{cc|cccc|l}
            \toprule[0.8pt]
             \multicolumn{2}{c|}{Task} & \multicolumn{4}{c|}{Optic Cup/Disc Segmentation} & \multirow{2}{*}{Avg.} \\
             \cmidrule[0.3pt]{1-6}
             $\textbf{RAM}_{\text{Aug}}$ & \textbf{DSIR} & Domain 1 & Domain 2 & Domain 3 & Domain 4 & \\
             \midrule[0.3pt]
             - & - & 81.44/95.52 & 77.20/87.96 & 85.11/94.56 & 72.30/90.97 & 85.63 \\
             \checkmark & - & 83.06/94.86 & 78.09/89.04 & 86.73/\textbf{95.01} & 82.28/92.89 & 87.75 \\
             - & \checkmark & 83.76/95.31 & 77.43/88.07 & 85.84/94.19 & 81.58/91.48 & 87.21 \\
             \checkmark & \checkmark & \textbf{85.48}/\textbf{95.75} & \textbf{78.82}/\textbf{89.43} & \textbf{87.44}/94.67 & \textbf{85.84}/\textbf{94.10} & \textbf{88.94} \\
             \bottomrule[0.8pt]
        \end{tabular}
    }
    \label{fundus_ablation}
\end{table}

\begin{table}[t]
    \centering
    \caption{Ablation Study of key components in our method on \textbf{Prostate} Segmentation Task (\%). We mark the top results in \textbf{bold}.}
    \scriptsize
    \setlength{\tabcolsep}{0.5mm}{
        \begin{tabular}{cc|cccccc|l}
            \toprule[0.8pt]
             \multicolumn{2}{c|}{Task} & \multicolumn{6}{c|}{Prostate Segmentation} & \multirow{2}{*}{Avg.} \\
             \cmidrule[0.3pt]{1-8}
             $\textbf{RAM}_{\text{Aug}}$ & \textbf{DSIR} & Domain 1 & Domain 2 & Domain 3 & Domain 4 & Domain 5 & Domain 6 & \\
             \midrule[0.3pt]
             - & - & 85.30 & 87.56 & 82.33 & 87.37 & 80.49 & 81.40 & 84.04 \\
             \checkmark & - & 87.28 & 89.94 & 85.45 & 87.86 & 86.17 & 86.94 & 87.27 \\
             - & \checkmark & 86.57 & 88.04 & 83.19 & 87.42 & 82.08 & 83.14 & 85.07 \\
             \checkmark & \checkmark & \textbf{87.56} & \textbf{90.20} & \textbf{86.92} & \textbf{88.72} & \textbf{87.17} & \textbf{87.93} & \textbf{88.08} \\
             \bottomrule[0.8pt]
        \end{tabular}
    }
    \label{prostate_ablation}
\end{table}

As aforementioned, during the training process, we employed a semantic consistency loss as a supervision signal to make the model resistant to domain shift. In Fig.~\ref{ablation_consist}, we investigate the effectiveness of our semantic consistency loss on \textbf{Fundus} and \textbf{Prostate} datasets. We observe that without the semantic consistency loss, all of the results degenerate on both datasets. This indicates that the semantic consistency loss do help improve the generalization performance of our model which means our model can be more robust to domain shift.

\begin{table}[!htbp]
    \centering
    \caption{Dice coefficient of different consistency loss on \textbf{Fundus} segmentation task (\%). We mark the top results in \textbf{bold}.}
    \scriptsize
    \setlength{\tabcolsep}{2mm}{
        \begin{tabular}{c|cccc|c}
            \toprule[0.8pt]
            Task & \multicolumn{4}{c|}{Optic Cup/Disc Segmentation} & \multirow{2}{*}{Avg.} \\
            \cmidrule[0.3pt]{1-5}
            Unseen Site & Domain 1 & Domain 2 & Domain 3 & Domain 4 & \\
            \midrule[0.3pt]
            MSE & 85.45/95.13 & 77.96/89.14 & 86.73/\textbf{94.76} & \textbf{85.93}/\textbf{94.16} & 88.65 \\
            JS-Div & 85.04/94.91 & 78.02/88.27 & 86.32/93.91 & 85.14/93.87 & 88.19 \\
            KL-Div & \textbf{85.48}/\textbf{95.75} & \textbf{78.82}/\textbf{89.43} & \textbf{87.44}/94.67 & 85.84/94.10 & \textbf{88.94} \\
            \bottomrule[0.8pt]
        \end{tabular}
    }
    \label{ablation_kl_fun}
\end{table}

Moreover, we experiment different types of consistency loss on \textbf{Fundus} dataset. In Table~\ref{ablation_kl_fun}, we show the results of different kinds of consistency loss. Except for KL-divergence (KL-Div), we also employ mean squared error (MSE) and Jensen–Shannon divergence (JS-Div). We observe that using different consistency loss will not affect the overall results of our method much, which means our method is robust to different types of consistency loss.

\section{Conclusion}
We present a novel generalizable medical image segmentation method for fundus and prostate image segmentation. To combat with overfitting in DG segmentation, we introduce random amplitude mixup (RAM) module to synthesize images with different domain style. We utilize the synthetic images as data augmentation to train the segmentation model and propose a self-supervised domain-specific image restoration (DSIR) module to recover the original images from synthetic images. Moreover, to further make the model resistant to domain shift and learn more domain invariant feature representation, we employ a semantic consistency loss in our training process. Our experimental results and ablation analysis indicate that all of the proposed components can help regularize the model and improve generalization performance on unseen target domains.

\noindent\textbf{Acknowledgements.} This work was supported by NSFC Major Program (62192783), CAAI-Huawei MindSpore Project (CAAIXSJLJJ-2021-042A), China Postdoctoral Science Foundation Project (2021M690609), Jiangsu Natural Science Foundation Project (BK20210224), and CCF-Lenovo Bule Ocean Research Fund.
\clearpage
%
%
\bibliographystyle{splncs04}
\bibliography{egbib}

\end{sloppypar}
\end{document}